\documentclass{article}
\usepackage{spconf,amsmath,epsfig}
\usepackage{mathrsfs}
\usepackage{amsfonts}
\usepackage{subfigure}
\usepackage{algorithmic}
\usepackage{listings}
\usepackage{tabularx}
\usepackage{verbatim}
\usepackage{pdflscape}
\usepackage{mathtools}
\usepackage{epstopdf}

\usepackage[belowskip=-15pt,aboveskip=0pt]{caption}
\usepackage[font=footnotesize,labelfont=bf]{caption}

\begin{document}\sloppy

\def\x{{\mathbf x}}
\def\L{{\cal L}}

\title{Multiresolution Match Kernels for Gesture Video Classification\vspace{-1ex}}
%

\name{Hemanth Venkateswara, Vineeth N. Balasubramanian, Prasanth Lade, Sethuraman Panchanathan\vspace{-2ex}}
\address{Center for Cognitive Ubiquitous Computing(CUbiC), Arizona State University \\
\texttt{\{hemanthv, vineeth.nb, prasanthl, panch\}@asu.edu}}
%
%
%

\begin{footnotesize}
\maketitle
\end{footnotesize}

\begin{abstract}
The emergence of depth imaging technologies like the Microsoft Kinect has renewed interest in computational methods for gesture classification based on videos. For several years now, researchers have used the Bag-of-Features (BoF) as a primary method for generation of feature vectors from video data for recognition of gestures. However, the BoF method is a coarse representation of the information in a video, which often leads to poor similarity measures between videos. Besides, when features extracted from different spatio-temporal locations in the video are pooled to create histogram vectors in the BoF method, there is an intrinsic loss of their original locations in space and time. In this paper, we propose a new Multiresolution Match Kernel (MMK) for video classification, which can be considered as a generalization of the BoF method. We apply this procedure to hand gesture classification based on RGB-D videos of the American Sign Language(ASL) hand gestures and our results show promise and usefulness of this new method.
\end{abstract}
\section{\vspace{-1ex}Introduction}
\label{sec_intro}
\vspace{-1ex}With the emergence of depth sensors like the Microsoft Kinect, there has been a renewed interest in the analysis of gestures using both color (or grayscale) and depth information. A standard procedure to estimate such features from videos, is to extract points of interest in the video (such as SIFT~\cite{lowe2004distinctive}) and subsequently obtain feature descriptors using these interest points~\cite{weinland2011survey}, which are then used to train a classifier. One of the most common methods used by researchers today in this regard is the Bag-of-Features (BoF) model~\cite{laptev2005space}.\\
\indent The BoF model is applied upon videos that are represented by a set of interest point based descriptors. A dictionary of codewords that represents the set of descriptors from all the videos is estimated. The dictionary is usually the centers of descriptor clusters. Each descriptor in a video is then mapped to the closest codeword in the dictionary, resulting in a histogram over the codewords. Two descriptors are similar if they map to the same codeword. The similarity between two videos is approximated by the similarity between their descriptor histograms. The histogram similarity is a coarse measure due to the binarization of the belongingness of a descriptor to a cluster.
Recently, Bo et al.~\cite{bo2009efficient} applied kernel methods to measure descriptor similarity and devised an efficient method to use these kernels to evaluate image similarity for object recognition in images. In this paper, we show how the concept of efficient kernels can be extended to estimate a kernel-based similarity between videos.\\
\indent Further, the BoF procedure has an additional drawback - it inherently leads to a global comparison of the similarity between two videos. Pooling descriptors from different $(x,y,t)$ locations in a video into a histogram leads to loss of spatio-temporal information. Two unique gestures that have similar movements but differ in the order they are performed, will generate similar descriptors, though at different $(x,y,t)$ locations. Their histograms, from the BoF model, will however be similar. These gestures will get the same class label under any classification based on these histograms. To overcome this drawback, in this paper, we further extend the efficient kernels for videos to a spatio-temporal pyramid-based multiresolution model.\\
\indent In the proposed method, which we call Multiresolution Match Kernels (MMK), we bring together the concepts of efficient match kernels~\cite{bo2009efficient} and spatio-temporal pyramids~\cite{lazebnik2006beyond} for gesture video classification. We introduce a multiresolution spatio-temporal feature representation of a video. This representation entails a histogram based similarity measure at multiple resolutions. By replacing the binary histogram with a kernel function, we obtain a fine-grained rich similarity measure.  However, the computational cost of estimating kernels at multiple resolutions is high. We then derive a cost-effective formulation for MMK, motivated by~\cite{bo2009efficient}, at multiple resolutions. We apply the MMK to classify American Sign Language(ASL) hand gestures obtained using a Microsoft Kinect. The MSRGesture3D dataset~\cite{kurakin2012real}, has depth videos of the hands performing $12$ gestures across $10$ users. We demonstrate how MMK improves recognition accuracies with increased pyramid resolution and show how it performs better than the existing techniques.\\
\indent The remainder of this paper is organized as follows. Section \ref{sec_bgd} discusses earlier related work, followed by a description of the proposed MMK method in Section \ref{sec_mmk}. Section \ref{sec_expts} presents our experimental results, followed by our conclusions in Section \ref{sec_conc}.
\section{\vspace{-1ex}Related Work}
\label{sec_bgd}
\vspace{-1ex}Gesture recognition has been studied extensively for several decades now, and has been applied in various fields ranging from social robotics to sign language recognition. Until recently, gesture recognition was mostly studied using color(RGB) (or grayscale) videos. The introduction of the Kinect has renewed interest in gesture recognition using color and depth(RGB-D) videos. Weinland at al.~\cite{weinland2011survey} provide a survey of gesture recognition techniques with a unique classification of the different procedures used. One of the foremost techniques for video feature representation, the BoF, was popularized by Laptev et al.~\cite{laptev2005space}, where Histogram of Oriented Gradients (HOG) and Histogram of Optical Flow (HOF) based descriptors were estimated from local neighborhoods of spatio-temporal interest points in a video. A BoF procedure was then applied to estimate histogram vectors that represented the videos.\\
\indent Grauman and Darrell~\cite{grauman2005pyramid} extended the standard BoF by introducing a spatial pyramid for object recognition. The inclusion of spatial information into a pyramid based BoF, was formalized by Lazebnik et al.~\cite{lazebnik2006beyond}. Spatio-temporal pyramids for videos have been implemented in~\cite{xu2007visual, choi2008spatio} which use temporal pyramid alignment and histogram intersection respectively, to estimate video similarities. In this paper, we improve upon the pyramid similarity measure by using Multiresolution Match Kernels (MMK). We use the idea of efficient kernels~\cite{bo2009efficient} to overcome the coarse binning shortcoming of the BoF by efficiently estimating kernel based similarity between descriptors across videos. We now describe our method.

\section{\vspace{-1ex}Multiresolution Match Kernels}
\label{sec_mmk}
\vspace{-1ex}In the BoF procedure, every gesture video is represented as a histogram. The data being binned is the set of feature descriptors from the video and the bin centers are a set of dictionary vectors. Let $\mathbf{X}\!=\!\{\mathbf{x}_1,\mathbf{x}_2,\hdots,\mathbf{x}_m\} $ represent the feature descriptors in a video, and $\mathbf{V}\! =\!\{\mathbf{v}_1,\mathbf{v}_2,\hdots,\mathbf{v}_D\}$, be a set of $D$ dictionary vectors. A feature descriptor is quantized into a binary vector $\mu(\mathbf{x}) \!=\![\mu_1(\mathbf{x}),\mu_2(\mathbf{x}),\hdots,\mu_D(\mathbf{x})]^\top$. $\mu_i(\mathbf{x})$ is 1 if $\mathbf{x} \in R(\mathbf{v}_i)$ and 0 otherwise, where $R(\mathbf{v}_i) \!=\!\{\mathbf{x} : \|\mathbf{x}-\mathbf{v}_i\| \leq \|\mathbf{x}-\mathbf{v}\|, \forall \mathbf{v} \in \mathbf{V}\}$ (as in~\cite{bo2009efficient}). The histogram feature vector for $\mathbf{X}$ is then given by $\bar{\mu}(\mathbf{X})\!=\!\frac{1}{|\mathbf{X}|}\sum_{\mathbf{x}\in\mathbf{X}}\mu(\mathbf{x}).$ A linear kernel measuring video similarity between two videos $\mathbf{X}$ and $\mathbf{Y}$ is a dot product over the histogram vectors, given by $K(\mathbf{X},\mathbf{Y}) = \frac{1}{|\mathbf{X}||\mathbf{Y}|}\sum_{\mathbf{x}\in\mathbf{X}}\sum_{\mathbf{y}\in\mathbf{Y}}\mu(\mathbf{x})^\top\mu(\mathbf{y})$. This is the standard BoF similarity measure for two videos.
$\mu(\mathbf{x})^\top\mu(\mathbf{y})$ can be expressed as a positive definite function $\delta(\mathbf{x},\mathbf{y})$, where $\delta(\mathbf{x},\mathbf{y})$ is 1 if $\{\mathbf{x},\mathbf{y}\}\subset R(\mathbf{v}_i), \exists{i}\in\{1,\hdots,D\}$, and 0 otherwise.\\
\begin{figure}[hbtp]
\begin{center}
\subfigure{
\includegraphics[width=1.0in, height=0.9in]{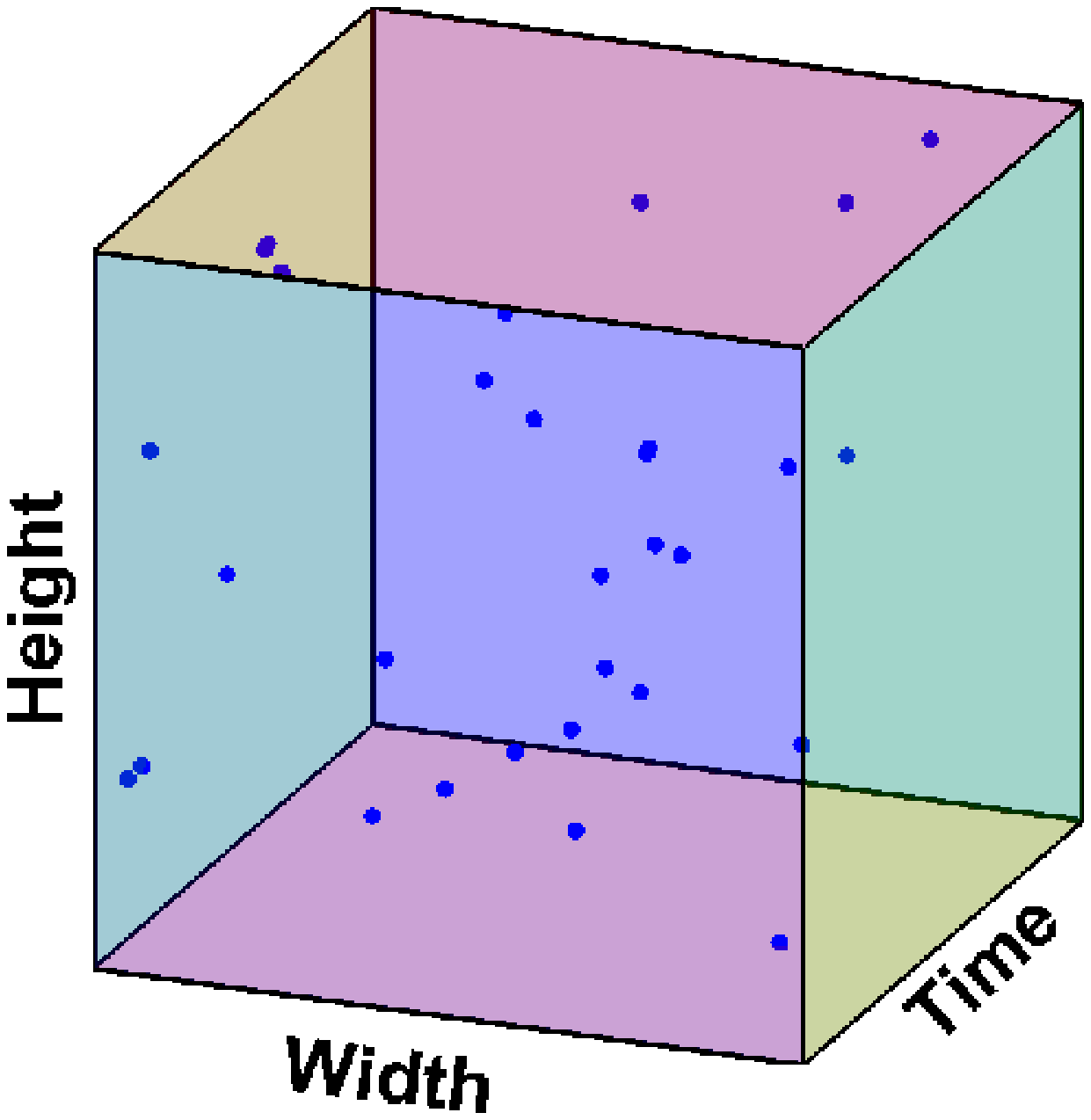}}
\subfigure{
\includegraphics[width=1.0in, height=0.9in]{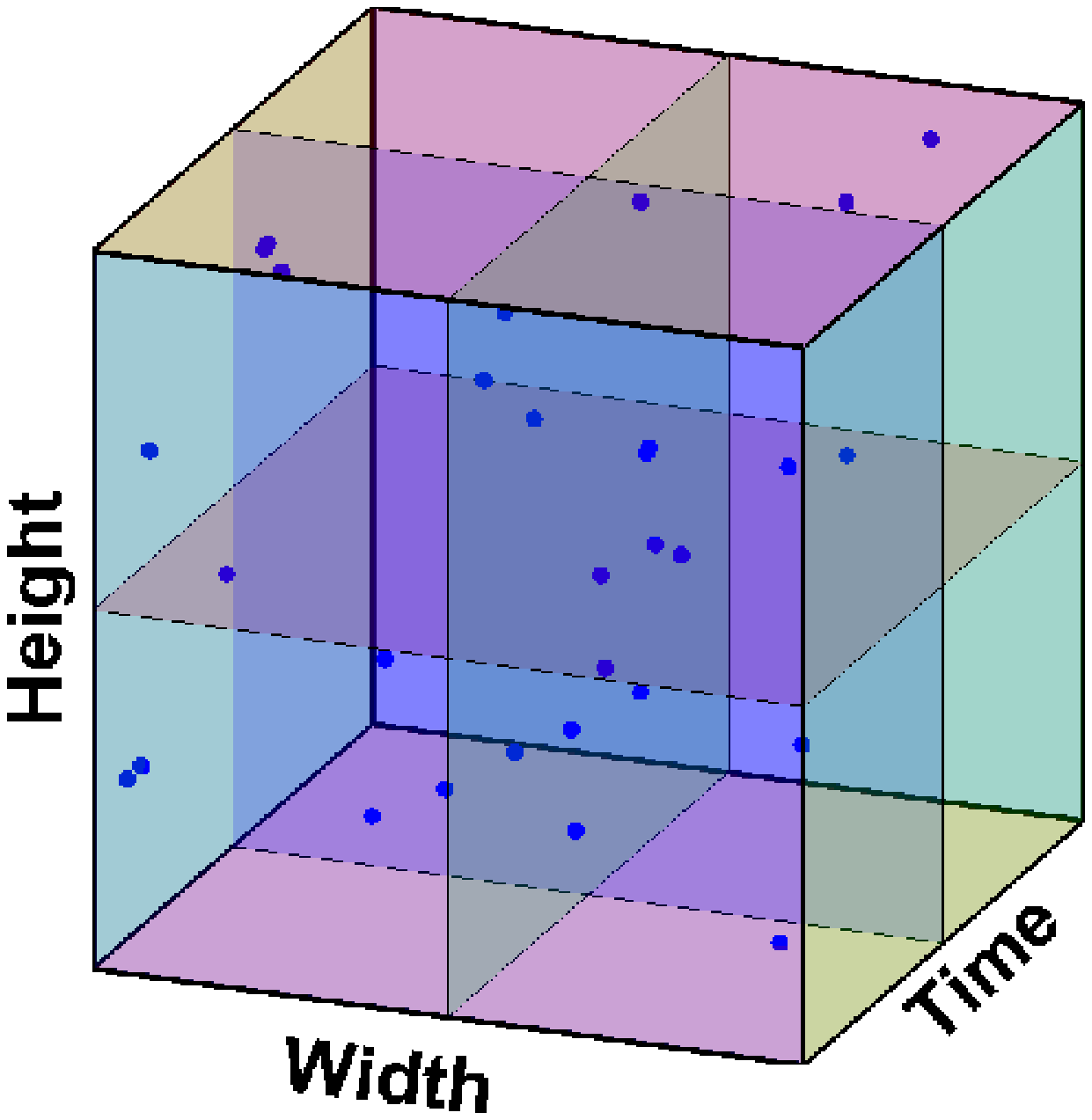}}
\subfigure{
\includegraphics[width=1.0in, height=0.9in]{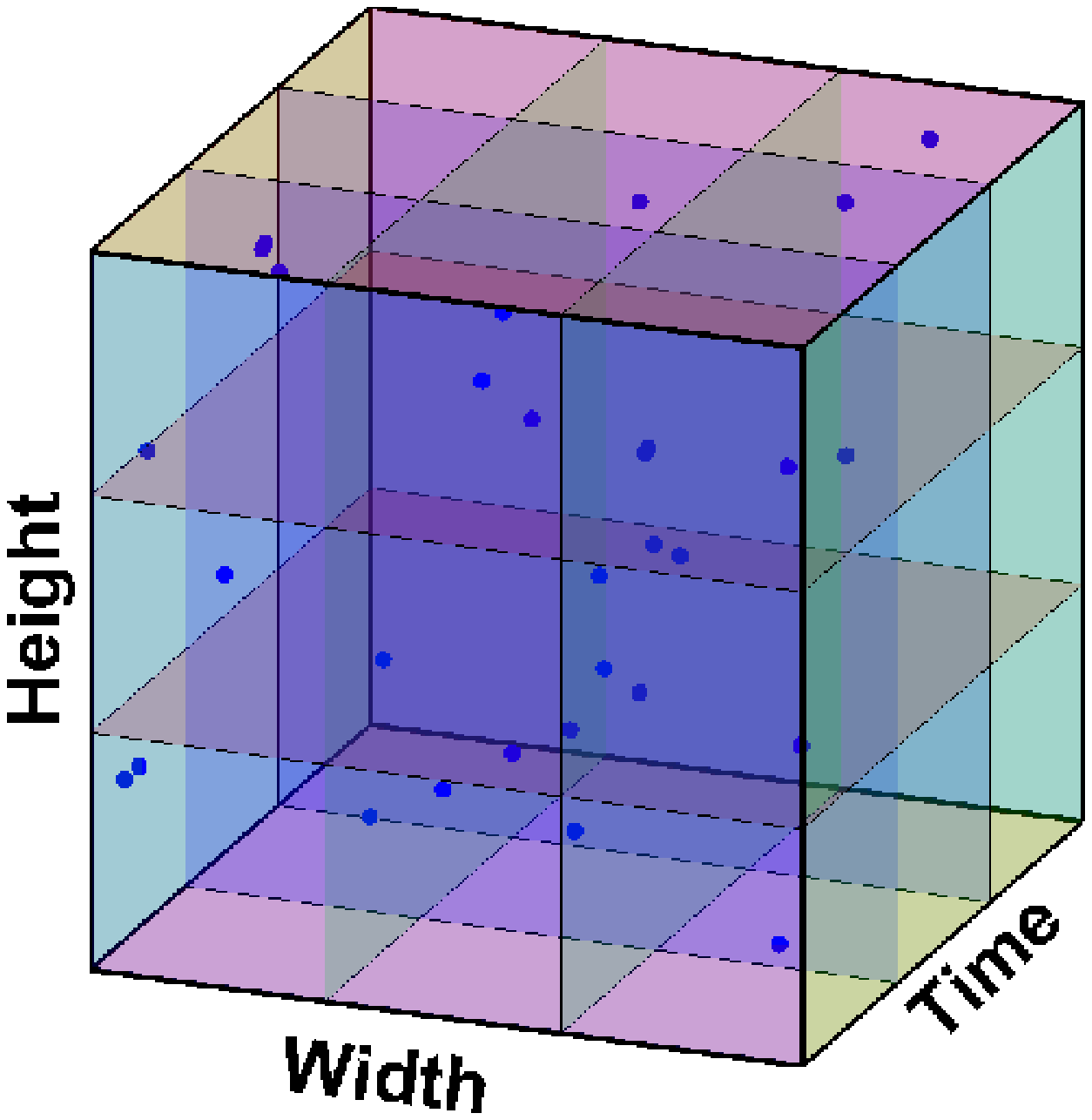}}
\caption{Splitting a video into voxels at multiple levels: (\textbf{left}) Level-1, (\textbf{center}) Level-2, (\textbf{right}) Level-3.}
\label{cubesFig}
\end{center}
\end{figure}
\indent As mentioned earlier, the BoF representation of $\mathbf{X}$ does not retain information about the spatio-temporal locations of the original feature descriptors. To overcome this drawback, we propose a multiresolution spatio-temporal pyramid representation for videos. A video is represented as a single cuboid of dimensions $[Height \times Width \times Time]$. We split the video cuboid into smaller cuboids called voxels. At level $L=1$ the video is made up of only 1 voxel. At Level $L=2$ the video is split into $2^3\!=\!8$ voxels. At level $L$ the video has $L^3$ voxels. A spatio-temporal pyramid at level $L$ consists of a sequence of all the voxels generated from levels $[1,\hdots,L]$. The number of voxels for a spatio-temporal pyramid $L=3$ is therefore $1^3+2^3+3^3=36$. Figure.~\ref{cubesFig}, represents video voxels for $3$ levels. The dots refer to the interest points at $(x,y,t)$ that fall within different voxels at different levels. We represent the descriptors for a given level as an ordered partition over the descriptor set $\mathbf{X} \!=\!\{\mathbf{x}_1,\mathbf{x}_2,\hdots,\mathbf{x}_m\} $ in the video. An ordered partition for level $i$ is represented as $\mathbf{X}^i\!=\!\{\mathbf{X}^{i,1}, \mathbf{X}^{i,2},\hdots,\mathbf{X}^{i,T_i}\}$ where $T_i$ is the number of partitions(voxels) at level $i$, with $T_i\!=\!i^3$. For $\mathbf{x} \in \mathbf{X}^{i,j}$, where $\mathbf{X}^{i,j} \subseteq \mathbf{X}$ is the set of descriptors at level $i$ and voxel $j$, the binary vector representation is given as $\mu(\mathbf{x}) \!=\![\mu_1(\mathbf{x}),\mu_2(\mathbf{x}),\hdots,\mu_D(\mathbf{x})]^\top$. For the sake of simplicity, we deploy the same dictionary $\mathbf{V}$ for quantization. The histogram of descriptors for voxel $(i,j)$ is represented as $\bar{\mu}(\mathbf{X}^{i,j})\!=\!\frac{1}{|\mathbf{X}^{i,j}|}\sum_{\mathbf{x}\in\mathbf{X}^{i,j}}\mu(\mathbf{x})$.  The similarity measure between two corresponding voxels (both labeled $(i,j)$) from two videos $\mathbf{X}$ and $\mathbf{Y}$ is given as $\bar{\mu}(\mathbf{X}^{i,j})^\top\bar{\mu}(\mathbf{Y}^{i,j})\!=\!\frac{1}{|\mathbf{X}^{i,j}||\mathbf{Y}^{i,j}|}\sum_{\mathbf{x}\in\mathbf{X}^{i,j}}\sum_{\mathbf{y}\in\mathbf{Y}^{i,j}}\mu(\mathbf{x})^\top\mu(\mathbf{y})$. For a given level, the multiresolution BoF similarity between two videos $\mathbf{X}$ and $\mathbf{Y}$, is the sum of the similarities of histogram vectors from corresponding voxels in videos $\mathbf{X}$ and $\mathbf{Y}$. A linear kernel measuring the multiresolution BoF similarity between two videos is now the sum over multiple dot products spanning over corresponding voxels at all the levels. It is given by:
\begin{equation}
K(\mathbf{X},\mathbf{Y}) = \sum_{i=1}^{L}\sum_{j=1}^{T_i}\frac{1}{|\mathbf{X}^{i,j}||\mathbf{Y}^{i,j}|}\sum_{\mathbf{x}\in\mathbf{X}^{i,j}}\sum_{\mathbf{y}\in\mathbf{Y}^{i,j}}\mu(\mathbf{x})^\top\mu(\mathbf{y})
\label{eq2}
\end{equation}
Unlike in the standard BoF, the kernel in Equation~(\ref{eq2}) measures the similarity between two videos at multiple resolutions, maintaining locality in spatial and temporal dimensions. Greater the number of levels, the more finer the comparison between the videos. When $L\!=\!1$, Equation~(\ref{eq2}) reduces to the standard BoF similarity measure used for videos as mentioned at the beginning.\\
\indent We now introduce the Multiresolution Match Kernels. The dot product $\mu(\mathbf{x})^\top\mu(\mathbf{y})$ in Equation~(\ref{eq2}) can be replaced by the positive definite function $\delta(\mathbf{x},\mathbf{y})$, as discussed previously. Since $\delta(\mathbf{x},\mathbf{y})$ is a coarsely quantized estimate of the similarity between $\mathbf{x}$ and $\mathbf{y}$, we replace it by a finite-dimensional kernel function $k(\mathbf{x},\mathbf{y})$, to estimate the similarity between $\mathbf{x}$ and $\mathbf{y}$~\cite{bo2009efficient}. This modifies Equation~(\ref{eq2}) to:
\begin{equation}
K(\mathbf{X},\mathbf{Y}) =\sum_{i=1}^{L}\sum_{j=1}^{T_i}\frac{1}{|\mathbf{X}^{i,j}||\mathbf{Y}^{i,j}|}\sum_{\mathbf{x}\in\mathbf{X}^{i,j}}\sum_{\mathbf{y}\in\mathbf{Y}^{i,j}}k(\mathbf{x},\mathbf{y}) 
\label{eq3}
\end{equation}
Equation~(\ref{eq3}) can be viewed as a sum of kernels for the same descriptors $(\mathbf{x},\mathbf{y})$ over multiple levels.\\
\indent Let us define $k_{i,j}(\mathbf{X},\mathbf{Y}) \!:=\! \frac{1}{|\mathbf{X}^{i,j}||\mathbf{Y}^{i,j}|}\sum_{\mathbf{x}\in\mathbf{X}}\sum_{\mathbf{y}\in\mathbf{Y}}k(\mathbf{x},\mathbf{y})$ where $k(\mathbf{x},\mathbf{y})=0$ for ${x}\notin\mathbf{X}^{i,j}$ or ${y}\notin\mathbf{Y}^{i,j}$. Equation~(\ref{eq3}) can then be written as $K(\mathbf{X},\mathbf{Y})\!=\!\sum_{i=1}^{L}\sum_{j=1}^{T_i}\alpha_{i}k_{i,j}(\mathbf{X},\mathbf{Y})$, where $\alpha_{i}$ is a weight for the kernel at Level $i$. We intend to give more weight to kernel similarities at finer resolutions(higher levels). The kernel $K(\mathbf{X},\mathbf{Y})$ can be viewed as a linear combination of kernel functions across levels and voxels, which we call a Multiresolution Match Kernel (MMK). The computational cost of estimating MMKs through Equation~(\ref{eq3}) is prohibitively high, viz. $O(L^3m^2d)$ and $O(n^2L^3m^2d)$ to estimate the kernel similarity between two videos and to estimate the kernel gram matrix for $n$ videos, respectively, where $m$ is the average number of descriptors in a video and $d$ is the dimensionality of the descriptors. To overcome this computational overhead we estimate the feature mapping induced by the finite dimensional kernel $K(\mathbf{X},\mathbf{Y})$ and work in the mapped feature space as in~\cite{bo2009efficient}. We now proceed to estimate the feature representations $\bar{\phi}(\mathbf{X})$ induced by the kernel $K(\mathbf{X},\mathbf{Y}) = \bar{\phi}(\mathbf{X})^\top\bar{\phi}(\mathbf{Y})$.\\
\indent Gonen and Alpaydin stated in~\cite{gonen2011multiple} that a linear combination of kernels directly corresponds to a concatenation in the feature space. Therefore, the MMK induces a feature space that is a concatenation of the feature spaces induced by the individual kernels $k_{i,j}(\mathbf{X},\mathbf{Y})$. That is, if $k_{i,j}(\mathbf{X},\mathbf{Y})$ induces a feature space $\bar{\phi}_{i,j}(\mathbf{X})$, where $k_{i,j}(\mathbf{X},\mathbf{Y}) \!=\! \bar{\phi}_{i,j}(\mathbf{X})^\top\bar{\phi}_{i,j}(\mathbf{Y})$, the feature space induced by the MMK, $K(\mathbf{X},\mathbf{Y})$ is $\bar{\phi}(\mathbf{X}) \!=\! [\bar{\phi}_{1,1}(\mathbf{X})^\top,\hdots,\bar{\phi}_{i,j}(\mathbf{X})^\top,\hdots,\bar{\phi}_{L,T_{L}}(\mathbf{X})^\top]^\top$, with $T_L \!=\! L^3$. We now need to estimate the feature map $\bar{\phi}_{i,j}(\mathbf{X})$ induced by the kernel $k_{i,j}(\mathbf{X},\mathbf{Y})$. This involves the estimation of the feature map $\phi(\mathbf{x})$ induced by $k(\mathbf{x},\mathbf{y})=\phi(\mathbf{x})^\top\phi(\mathbf{y})$. We estimate $\phi(\mathbf{x})$ as in~\cite{bo2009efficient}. Given $\phi(\mathbf{x})$, $\bar{\phi}_{i,j}(\mathbf{X})$ is estimated as the mean of all the $\phi(\mathbf{x})$ that belong to voxel $(i,j)$. $\bar{\phi}(\mathbf{X})$ is then equal to $[\bar{\phi}_{1,1}(\mathbf{X})^\top,\hdots,\bar{\phi}_{i,j}(\mathbf{X})^\top,\hdots,\bar{\phi}_{L,T_{L}}(\mathbf{X})^\top]^\top$, with $T_L \!=\! L^3$.\\
\indent In estimating $\phi(\mathbf{x})$, we determine a set of $D$ basis vectors(dictionary) that maps the space of all $\mathbf{x}\in\mathbf{X}$,$\forall\mathbf{X}$. We denote this by a set of $D$ vectors, $\bar{\mathbf{H}}\!=\!\{\mathbf{z}_1,\mathbf{z}_2,\hdots,\mathbf{z}_D\}$. We estimate $\bar{\mathbf{H}}$ using K-means clustering with $D$ centers. $k(\mathbf{x},\mathbf{y})$ is given by $k(\mathbf{x},\mathbf{y}) \!=\! \mathbf{k}_Z(\mathbf{x})^{\top}\mathbf{K}_{ZZ}^{-1}\mathbf{k}_Z(\mathbf{y})$, where $\mathbf{k}_Z(\mathbf{x})$ is a $D\times1$ vector with $\{\mathbf{k}_Z\}_i\!=\!k'(\mathbf{x},\mathbf{z}_i)$. $\{\mathbf{K}_{ZZ}\}$ is a $D\times D$ matrix with $\{\mathbf{K}_{ZZ}\}_{i,j}\!=\!k'(\mathbf{z}_i,\mathbf{z}_j)$ for some positive definite kernel $k'(,)$ (for our experiments we use a RBF kernel). For $\mathbf{G}^\top\mathbf{G}=\mathbf{K}_{ZZ}^{-1}$, the feature map $\phi(\mathbf{x})\!=\!\mathbf{Gk}_Z(\mathbf{x})$. The feature map for voxel $(i,j)$ is then, $\bar{\phi}_{i,j}(\mathbf{X})\!=\!\frac{1}{|\mathbf{X}^{i,j}|}\mathbf{G}[\sum_{\mathbf{x}\in\mathbf{X}^{i,j}}\mathbf{k}_Z(\mathbf{x})]$. Please refer~\cite{bo2009efficient} for more details.\\
\indent The computational complexity of estimating $\bar{\phi}_{i,j}(\mathbf{X})$ is now $O(mDd + D^2)$. $\bar{\phi}(\mathbf{X})$ is obtained by concatenating the $\bar{\phi}_{i,j}(\mathbf{X})$. $\bar{\phi}(\mathbf{X})$ is the new multiresolution spatio-temporal representation of a video $\mathbf{X}$. If dictionary length is $D$, and the pyramid has $L$ levels, $\bar{\phi}(\mathbf{X})$ is of $D[\frac{L(L+1)}{2}]^2$ dimensions (e.g., for a dictionary of length $D\!=\!10$, and spatio-temporal pyramid level $L\!=\!3$, the total length of the feature map for a MMK is 360 - a level $L=3$ pyramid includes levels $L=2$ and $L=1$ also). This is a significant improvement in making the MMKs practical and efficient.

\section{\vspace{-1ex}Experiments and Results}
\label{sec_expts}
\begin{figure}
\centering
\includegraphics[width=3.0in, height=1.0in]{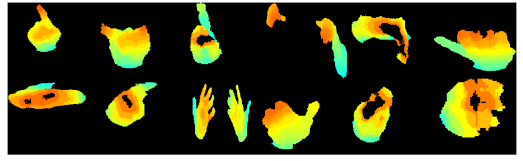}
\caption{Different gestures from the MSRGesture3D dataset~\cite{kurakin2012real}.}
\label{allGestures}
\end{figure}

\vspace{-1ex}The MSRGesture3D dataset~\cite{kurakin2012real} consists of $12$ depth gesture videos depicting letters and words from the American Sign Language, viz. \emph{bathroom, blue, finish, green, hungry, milk, past, pig, store, where, j, z}. It is performed by $10$ users and has $336$ examples. Figure.~\ref{allGestures} depicts a snapshot from an example of each gesture across different users. In this work, we extract interest points using SIFT~\cite{lowe2004distinctive}, from every frame in the depth video. Every frame is a gray scale image where the pixel value denotes the depth of the pixel in the image. The $(x,y,t)$ coordinates of these points are used to group them into voxels ($x,y$ are pixel coordinates, $t$ is the frame number). For this study we experiment with only 3 levels for the spatio-temporal pyramid. We apply MMK upon three kinds of feature descriptors (SIFT itself, Local Binary Patterns(LBP) and HOG) extracted from the aforementioned interest points to validate the performance of the proposed method.\\
\indent SIFT descriptors were estimated using a $16\times16$ window of pixels around each interest point. Each interest point yielded a $128$ dimension descriptor (as in~\cite{lowe2004distinctive}). LBP features~\cite{ojala2002multiresolution} were estimated from the same $16\times16$ pixel window around each interest point. For each pixel in the window, a $8$-bit LBP binary vector was estimated. The decimal values of the binary vectors were binned into a histogram with $59$ bins to result in a $59$ dimensional descriptor for each interest point. In case of HOG, for each of the pixels in the $16\times16$ pixel window around the interest point, the gradient orientations were estimated and binned into a histogram with $36$ bins. We apply the soft binning strategy from~\cite{bo2010kernel} to distribute the orientation about a set of bins. The feature descriptor for each interest point is a vector of $36$ components.
\begin{figure}
\centering
\includegraphics[width=2.5in, height=1.5in]{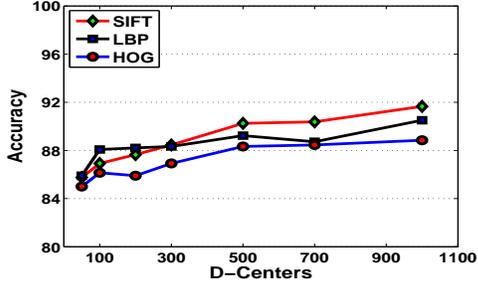}
\caption{Accuracies for multiple $D$ Values.}
\label{kMeansFig}
\end{figure}
\begin{figure}
\centering
\includegraphics[width=3.0in, height=2.4in]{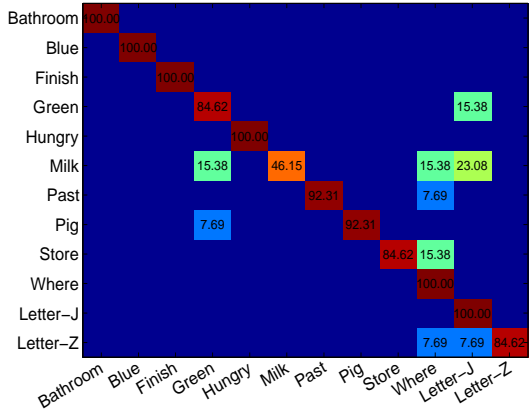}
\caption{Confusion matrix (SIFT features with MMK-3).}
\label{confusionMatrix}
\end{figure}
\indent For each kind of descriptor, we estimated the dictionary of $D$ codewords using K-means clustering. We empirically studied MMK with different $D$ values and estimated the best performance to be at $D=1000$ (see Figure.~\ref{kMeansFig}). Feature maps for level $i$ were weighted with $\alpha_{i}=1/2^{-i}$ giving higher weight to matches at finer resolutions. For $L=3$ spatio-temporal pyramid (which includes $L=1$ and $L=2$), the dimension of feature map $\bar{\phi}(\textbf{X})$, was $36000$. We use the LIBLINEAR classifier (as used by Bo et al.~\cite{bo2009efficient}) upon the features $\bar{\phi}(\textbf{X})$ for final gesture classification. Table~(\ref{table1}) shows the results of our experiments. Here, BoF is the standard Bag-of-Features method, while MMK-1, MMK-2 and MMK-3 are MMK with $L\!=\!1, L\!=\!2$ and $L\!=\!3$, respectively. These results were obtained using cross-validation, where $5$ subjects were used for training and $5$ subjects for testing. The obtained results were averaged over $5$ independent runs to address randomness bias.\\
\indent It is evident that the proposed MMK method outperforms BoF in these experiments. Also, the results show that the accuracies increase at higher resolutions of matching, corroborating the need for multiresolution match kernels. The confusion matrix for SIFT and MMK-3, shown in Figure.~\ref{confusionMatrix}, indicates that the gesture for \emph{Milk} was often misclassified. We believe that this is because the gesture involves a clenched fist whose descriptors are similar to descriptors from other gestures (like \emph{letter-J}). We will investigate ways to overcome this challenge in future work. Existing state-of-the-art on the MSRGesture3D dataset report accuracies $88.5\%$~\cite{kurakin2012real} and $87.7\%$~\cite{wang2012robust} using a leave-one-subject-out approach. We, hence, performed a study based on the leave-one-subject-out approach using the proposed MMK-3 method and obtained accuracies of $94.6\%$ (SIFT), $94.1\%$ (LBP) and $91.5\%$ (HOG), supporting our inference that the proposed method holds great promise for video-based classification tasks.
\begin{table}[ht]
\centering
\begin{tabular}{|c|c|c|c|c| }
\hline
& BoF & MMK-1 & MMK-2 & MMK-3\\ \hline
SIFT&71.28\% & 72.3\% & 87.56\% & 91.66\%  \\
LBP&66.15\% & 68.07\% & 85.64\% & 90.51\%  \\
HOG&74.10\% & 58.33\% & 74.87\% & 88.85\%  \\
\hline
\end{tabular}
\caption{Cross-validation accuracies(train $5$ subjects, test $5$ subjects).}
\label{table1}
\end{table}

\section{\vspace{-1ex}Conclusions}
\label{sec_conc}
\vspace{-1ex}We have proposed a Multiresolution Match Kernel framework for efficient and accurate estimation of video gesture similarity. Our method addresses two major drawbacks in the widely used Bag-of-Features method, viz., coarse similarity matching and spatio-temporal invariance. Our experiments on the MSRGesture3D dataset showed that the MMK performs significantly better than BoF across $3$ feature descriptors: SIFT, LBP and HOG. MMK provides a discriminative (multiresolution spatio-temporal) feature representation for videos with similar motion information that differ in the sequence of motion, where a standard BoF procedure would perform poorly. We plan to extend this method to other video-based classification tasks to study the generalizability of this approach in future work.
\footnotesize
\section{\vspace{-1ex}Acknowledgments}
\vspace{-1ex}This material is based upon work supported by the National Science Foundation under Grant No:1116360. Any opinions, findings, and conclusions or recommendations expressed in this material are those of the authors and do not necessarily reflect the views of the National Science Foundation.

\footnotesize
\vspace{-1ex}

\end{document}